\documentclass{IEEEcsmag}

\usepackage[colorlinks,urlcolor=black,linkcolor=blue,citecolor=black]{hyperref}
\usepackage{upmath}
\usepackage[ruled,vlined]{algorithm2e}
\usepackage{cite}
\usepackage{url}
\usepackage{multirow}
\usepackage{tcolorbox}

\jvol{XX}
\jnum{XX}
\paper{8}
\pubyear{2020}

\begin{document}


\title{Towards Hate Speech Detection at Large via Deep Generative Modeling}

\author{Tomer Wullach}
\affil{University of Haifa}

\author{Amir Adler}
\affil{Braude College of Engineering and\\ Massachusetts Institute of Technology}

\author{Einat Minkov}
\affil{University of Haifa}



\begin{abstract}
Hate speech detection is a critical problem in social media platforms, being often accused for enabling the spread of hatred and igniting physical violence. Hate speech detection requires overwhelming resources including high-performance computing for online posts and tweets monitoring as well as thousands of human experts for daily screening of suspected posts or tweets. Recently, Deep Learning (DL)-based solutions have been proposed for automatic detection of hate speech, using modest-sized training datasets of few thousands of hate speech sequences. While these methods perform well on the specific datasets, their ability to detect new hate speech sequences is limited and has not been investigated. Being a data-driven approach, it is well known that DL surpasses other methods whenever a scale-up in train dataset size and diversity is achieved. Therefore, we first present a dataset of 1 million realistic hate and non-hate sequences, produced by a deep generative language model. We further utilize the generated dataset to train a well-studied DL-based hate speech detector, and demonstrate consistent and significant performance improvements across five public hate speech datasets. Therefore, the proposed solution enables high sensitivity detection of a very large variety of hate speech sequences, paving the way to a fully automatic solution. 
\end{abstract}

\maketitle
\thispagestyle{empty}
\chapterinitial{Introduction} The rapidly growing usage of social media platforms and blogging websites has led to over 3.8 billion \cite{DIGITAL_2020} active social media users world wide (half of the world population) that use text as their prominent means of mutual communications. A fraction of the users use a language that expresses hatred towards a specific group of people, known as \textit{Hate Speech}  \cite{Twitter_2019,  Twitter_2020,Facebook_Hate_Speech_2020,TheVerge_Twitter_2020, Zuckerberg_2016}. Hate speech is mostly used against groups of common race, ethnicity, national origin, religious affiliation, sexual orientation, caste, gender, gender identity, and serious disease or disability. Hate speech can be expressed explicitly, for example: \emph{{"Refugees are the illness of the world and must be cured with antibiotics."}} or implicitly: \emph{{"Black women on London transport..."}}, which leads to a very large diversity of realistic hate speech sequences. Therefore, hate speech detection has become a critical problem for social media platforms, being often accused for enabling the spread of hatred and igniting physical violence in the real world. Hate speech detection requires overwhelming resources including high-performance computing for online posts and tweets monitoring. In addition, a portion of the suspected posts and tweets are sent for further screening by teams of thousands \cite{Facebook_Hatespeech} of human experts. It is therefore essential that automatic hate speech detection systems will identify in real-time hate speech sequences with a very high sensitivity, namely, having a low false-negative rate. Otherwise, a significant portion of the hate-speech will not be detected. \\
\indent Hate speech detection has been investigated extensively in recent years, and due to the outstanding success of Deep Learning (DL) \cite{Goodfellow-et-al-2016} in natural language processing \cite{8416973} it has become a leading approach for solving this problem.  Among the proposed DL architectures, the most well-studied ones such as \cite{zhang2018detecting}, were adopted from the related problem of sentiment analysis \cite{sentiment_analysis_A_survey}. The main reason for the success of DL is its \textit{Generalization} capability. Generalization \cite{Kawaguchi2017GeneralizationID} is defined as the ability of a classifier to perform well on unseen examples. In the supervised learning framework a classifier $\mathcal{C:X\rightarrow Y}$ is learned using a training dataset $\mathcal{S} = \{(x_1, y_1), . . . ,(x_N, y_N)\}$ of size $N$, where $x_i\in\mathcal{X}$ is a data example (i.e. a text sequence) and $y_i\in\mathcal{Y}$ is the corresponding label (i.e. hate or non-hate). Let $\mathcal{P(X,Y)}$ be the true distribution of the data, then the expected risk is defined by $\mathcal{R(C)}=E_{x,y \sim \mathcal{P(X,Y)}}[\mathcal{L(C}(x),y)]$, where $\mathcal{L}$ is a loss function that measures the misfit between the classifier output and the data label. The goal of DL is to find a classifier $\mathcal{C}$ that minimizes the expected risk, however, the expected risk cannot be computed since the true distribution is unavailable. Therefore, the empirical risk is minimized instead: $\mathcal{R_E(C)}=\frac{1}{N}\sum_{i=1}^{N}\mathcal{L(C}(x_i),y_i)$, which approximates the statistical expectation with an empirical mean. The \textit{Generalization Gap} is defined as the difference between the expected risk to the empirical risk: $\mathcal{R(C)} - \mathcal{R_E(C)}$. By using large and diverse datasets, DL has been shown to achieve a low generalization gap, where an  approximation of the expected risk is computed using the learned classifier and a held-out testing dataset $\mathcal{T} = \{(x_1, y_1), . . . ,(x_M, y_M)\}$ of size $M$, such that $\mathcal{S \cap T} = \emptyset.$ The implication of achieving a low generalization gap is good performance of the classifier on unseen data, namely, detecting accurately unseen hate speech.\\
\indent Existing hate speech detection solutions utilize public datasets with representative hate and non-hate sequences, collected from social media. These datsets contain only a few thousands of sequences, thus, with a limited capability to faithfully represent the large diversity of hate and non-hate sequences. Therefore, while good results are achieved for existing solutions, these are limited only to the scope of the utilized datasets, and have not been tested on a larger variety of hate/non-hate sequences. \\
\indent In this paper we present a novel approach for significantly increasing the generalization capabilities of hate-speech detection solutions. The performance of the proposed approach is evaluated using a well-studied DL-based detector, trained using five different public datasets.  First, we scale up significantly the training datasets in terms of size and diversity by utilizing the  GPT-2~\cite{radford2019language} deep generative language model. Each one of the five datasets was used to fine-tune the GPT-2 model, such that it generated $100,000$ hate and $100,000$ non-hate speech sequences, per dataset. Following the data generation stage, hate speech detection performance was evaluated by augmenting the baseline training sets with the generated data, demonstrating consistent and significant performance improvements of the DL detector, as compared to training using only the baseline (un-augmented) datsets.\\
\indent The contributions of this paper are three-fold: (i) the utilization of deep generative language modeling for approximating the probability distribution of hate speech and generating massive amounts of hate speech sequences. (ii) a comparative study of the achievable generalization of hate speech detection using existing hate speech datasets. (iii) a detailed demonstration of significant hate speech detection generalization improvements, by utilizing the generated data.\\
\indent The rest of this paper is organized as follows: an overview of hate speech detectors and datasets is provided in the related work section. The generation of hate and non-hate speech by the deep generative model is  detailed in the proposed approach section. Hate speech detection results and a study of the achievable generalization are provided in the performance evaluation section. The conclusions section summarizes the paper. 

\section{Related work}
\label{sec:related}

Researchers have proposed several approaches for hate speech detection over the recent years, ranging from classical learning methods, to modern deep learning. While these methods differ with respect to architecture and data representation, they share the need for a large, diverse, and high-quality data resource for training effective classification models. In the lack of such a resource, previous related works constructed datasets for this purpose, which were manually labeled, and therefore strictly limited in size. Below, we review existing hate speech detection approaches, and describe the datasets that have been used by researchers for training and evaluating such classifiers. We will leverage those labeled datasets which are publicly available for generating our MegaSpeech corpus. 

\subsection{Hate Speech Classifiers}
{\it Classical machine-learning methods} require data to be transformed into some pre-defined feature representation scheme, which should ideally capture meaningful and relevant information for classification purposes. Given textual data, features typically pertain to raw lexical information (surface word forms), considering individual words as well as word sequences (n-grams). Other features may encode  processed syntactic and semantic information. 
For example, Davidson {\it et al.}~\cite{davidson2017automated} represented tweets using word n-grams, having them weighted by word frequency (TF-IDF), part-of-speech and sentiment features. In addition, they encoded genre-specific features, denoting hashtag and user mentions, URLs, and the tweet's length. They then trained a logistic regression classifier to distinguish between tweets labeled as hate or offensive speech, or neither. In another work, a Support Vector Machine (SVM) was trained to predict whether a tweet was hateful or not using feature representations of word n-grams and typed word
dependencies~\cite{burnap2015cyber}.\\

{\it Deep Learning (DL) methods} can effectively exploit large-scale data, learning latent representations using multi-layered neural network architectures. Modern DL architectures of text processing generally consist of a word embedding layer, intended to capture generalized semantic meaning of words, mapping each word in the input sentence into a vector of low-dimension~\cite{mikolov2013distributed}. The following layers learn relevant latent feature representations, where the processed information is fed into a classification layer that predicts the label of the input sentence.

In a recent work, Founta {\it et al.}~\cite{founta2019unified} experimented with a popular DL architecture for text processing, classifying tweets into the categories of hate, abusive, or offensive speech, sarcasm and cybercullying. They transformed the input words into Glove word embeddings~\cite{pennington2014glove}. They then used a recurrent layer comprised of Gated Recurrent Units (GRU) for generating contextual and sequential word representations, having each word processed given the representations of previous words in the input sentence. Following a dropout layer (intended to prevent over-fitting), the final dense layer  outputs the probability that the sentence belongs to each of the target using a softmax activation function. Another popular choice of DL architecture for processing textual data is Convolution neural networks (CNN). Previously, Badjatiya {\it et al.}~\cite{badjatiya2017deep} used such an architecture for hate speech detection. CNN applies a filter over the input word representations, processing local features per fixed-size word subsequences. Hierarchical processing, which involves aggregation and down-sampling, consolidates the local features into global representations. These representations are finally fed into a fully-connected layer which predicts the probability distribution over all of the possible classes.

Another DL classification architecture presented by Zhang {\it et al.}~\cite{zhang2018detecting} is comprised of both convolution and recurrent  processing layers. This architecture has been shown to yield top-performance results on the task of hate speech classification, and is therefore the architecture of choice in our experiments. In brief, and as illustrated in Figure~\ref{fig:CNN_GRU}, this text classification network consists of a convolution layer applied to individual dimensions of the input word embeddings, which is down-sampled using a subsequent max-pooling layer. The following layer is recurrent (GRUs), producing hidden state representations per time step. Finally, a global max-pooling layer is applied, and a softmax layer produces a probability distribution over the target classes for the given input. Further implementation details are provided later in this paper.

\subsection{Hate speech datasets}
\label{ssec:datasets}

Table~\ref{tab:datasets} details the statistics of five public datasets of hate speech, which are all manually curated and of modest size. Below, we discuss the generation process and characteristics of these datasets. In this work, we will leverage these datasets for tuning an automatic language generation process, resulting the in a very large number of hate speech sequences, namely the Mega-speech resource.  

The typical process that has been traditionally applied by researchers and practitioners for constructing labeled datasets in general, and of hate speech in particular, involves the identification of authentic hate speech sentences, as well as counter (non-hate) examples. Twitter is often targeted as a source of relevant data; this (and other) public social media platform allows users to share their thoughts and interact with each other freely, making it a fertile ground for expressing all kinds of agendas, some of which may be racist or hateful. The initial retrieval of hate speech examples from Twitter is based on keyword matching, specifying terms that are strongly associated with hate.\footnote{e.g., \url{https://www.hatebase.org/}} Once candidate tweets are collected, they are assessed and labeled by human annotators into pre-specified categories. 
The manual annotation of the examples provides with high-quality ground truth labeled datasets, yet it is costly. Accordingly, the available datasets each include only a few thousands of labeled examples. Due to their size limit, and  biases involved in the dataset collection process, e.g., keyword selection and labeling guidelines, these datasets may be under-representative of the numerous forms and shapes in which hate speech may be manifested.\footnote{Another caveat of referring to authentic content within a dataset concerns the discontinued availability of the collected texts by the relevant provider over time. For example, tweets must be stored by their identifier number, where access to the tweet's content may be defined, imparing the dataset.} 

We now turn to describe the individual datasets in Table~\ref{tab:datasets} in more detail.  
The dataset due to Davidson {\it et al.}~\cite{davidson2017automated} (\textbf{DV}) includes tweets labeled by CrowdFlower\footnote{https://www.welcome.ai/crowdflower} workers into three categories: {\it hate speech}, {\it offensive}, or {\it neither}. For the purposes of this work, we only consider examples of the first and latter categories.  Waseem and Hovy~\cite{waseem2016hateful} created another dataset (\textbf{WS}), considering tweets of accounts which frequently used slurs and terms related to religious, sexual, gender and ethnic minorities; those tweets were manually labeled into the categories of {\it racism}, {\it sexism} or {\it neiter}. Again, we only consider the first and latter categories as examples of hate and non-hate, respectively. Another dataset was constructed by SemEval conference organizers~\cite{basile2019semeval} for the purpose of promoting hate detection (\textbf{SE}). They considered the historical posts of identified hateful Twitter users, narrowed down to tweets that included hateful terms, and had examples labeled by CrowdFlower workers.  We find that many of the tweets labeled as hateful in this dataset target women and immigrants. Founta {\it et al}~\cite{founta2018large} (\textbf{FN}) performed iterative sampling and exploration while having tweets annotated using crowdsourcing. Their resulting dataset is relatively large ($\sim$80K examples), and distinguishes between multiple flavors of offensive speech, namely {\it offensive}, {\it abusive}, {\it hateful}, {\it aggressive}, {\it cyber bullying}, {\it spam} and {\it none}. In order to maintain our focus on hate speech, we consider the labeled examples associated with the {\it hateful} and {\it none} categories. Finally, the dataset due to de Gilbert {\it et al.}~\cite{de-gibert-etal-2018-hate} (\textbf{WH}) was extracted from the extremist StormFront Internet forum,\footnote{https://en.wikipedia.org/wiki/Stormfront\_(website)}. This dataset aims to gauge hate in context, considering also cases where a sentence does not qualify as hate speech on its own, but is interpreted as hateful within a larger context comprised of several sentences.

\begin{table}[t]
\caption{Publicly available hate speech datasets, pertaining to examples labeled strictly as hate or non-hate.}
\label{tab:datasets}
\setlength\tabcolsep{0pt}
\begin{tabular*}{\columnwidth}{@{\extracolsep{\fill}} ll rr}
\toprule
     Dataset & Source & Hate & Non-Hate\\\hline\hline

{DV} \cite{davidson2017automated} & {Tweets} &  {1,404} & {7,875}\\
{FN} \cite{founta2018large} & {Tweets} &  {4,020} & {49,775}\\ 
{WS} \cite{waseem2016hateful}& {Tweets} & {1,965} & {3,797}\\
{WH} \cite{de-gibert-etal-2018-hate}& {Posts} &  {1,196} & {9,507}\\
{SE} \cite{basile2019semeval} & {Tweets} & {4,217} & {5,758}\\
\hline
\hline
\end{tabular*}
\end{table}

\section{The proposed approach: Scalable Hate Speech Detection via Data Generation}

\subsection{Hate Speech Data Generation}
\label{{ssec:approach}}

Ideally, DL-classifiers would be trained using large and diverse datasets, which more closely represent the true distribution of hate speech. We automatically develop large-scale datasets of hate speech text sequences and coutner examples from seed labeled examples using the GPT-2  model~\cite{radford2019language}. 

The Generative Pretrained Transformer 2 (GPT-2) is a language model, trained to predict, or "generate", the next token in a sequence given the previous tokens of that sequence in an unsupervised way. Being trained using 40GB of Web text, GPT-2 automatically generates text sequences of greater quality than ever seen before (CITE). We employed the large configuration of GPT-2 (764M parameters) in this work.\footnote{A yet larger GPT-2 model (1542M parameters) may yield even better text quality, but requires substantially larger computing resources.}

Our aim is to generate hateful rather than general text sequences, however. Concretely, we wish to generate hate (and, non-hate) speech sequences that: a) are perceived by humans as hate speech (or, non-hate) with high probability, b) introduce high language diversity, and c) exhibit high language fluency. In order to meet the first requirement, we exploit available labeled examples, tuning GPT-2 towards hate (and non-hate) speech generation. High diversity is achieved by leveraging the rich language model encoded into GPT-2, and by generating a very large number of candidate sequences. Last, we maintain high data quality by automatically evaluating the generated candidate sequences using a state-of-the-art text classifier, namely BERT~\cite{devlin2018bert}, where generated sequences that are evaluated as low-quality texts by BERT are automatically removed. Next, we formalize and discuss each of these components of our approach for hate speech generation.

\textbf{Domain fine-tuning} We wish to approximate the language distribution of a specific class, namely {\it hate speech} or {\it non-hate}, as opposed to generating random text sequences. We therefore {\it fine-tune} the pre-trained GPT-2 model, where we continue training the model from its distribution checkpoint, serving it with relevant text sequences. We randomly selected 80\% of the labeled examples per dataset and category for this purpose (see Table~\ref{tab:datasets}). (The remaining labeled examples are held out for evaluating hate detection performance in our experiments.)  
Let us denote the examples sampled from dataset $d_i$ that are labeled as {\it hate} and {\it non-hate} by $d^{i}_{hate}$ and $d^{i}_{non-hate}$, respectively. Accordingly, we fine-tune two distinct GPT-2 models per dataset $d_i$: 
\begin{itemize}
    \item $\mathbf{G}^i_{hate}$--fine-tuned using $d^{i}_{hate}$, and
    \item $\mathbf{G}^i_{non-hate}$--fine-tuned using $d^{i}_{non-hate}$.
\end{itemize}
Importantly, fine-tuning a separate GPT-2 model per dataset and class (yielding 10 models overall) is expected to bias the language generation process according to the topics and terms that are characteristic to that combination of dataset and class. Once text sequences are generated from each fine-tuned model, they are unified so as to obtain a diverse distribution of automatically generated text samples. 

\textbf{Candidate sequence Generation} Given the fine-tuned GPT-2 models, we generate a large number of sequences (600K) per model. Similarly to the labeled examples, we generate text sequences that are relatively short, limiting the sequence length to 30 tokens. We provide no prompt to the GPT-2 model, that is, text is generated unconditionally, starting from an empty string.\footnote{We set the generation parameters {\it top\_p} and {\it temperature} both to 0.9.} 

\textbf{Sequence selection}
One should not expect all of the sequences automatically generated by the fine-tuned GTP-2 models to be perceived as hateful (or non-hateful). We found it crucial to assess and filter the generated sequences by their quality and  relevance to the focus class.

We employ BERT~\cite{devlin2018bert},\footnote{We used the Bert-base model in our experiments} a powerful DL transformer-based model, which has been shown to achieve state-of-the-art results on many natural language processing tasks, for automatically classifying the generated sequences as hateful (or not). BERT, which has been trained using large amounts of text, projects given text sequences into contextual embedding representations, which are useful for text classification; adding a softmax layer on top of BERT, one may learn to predict target classes of interest by further training the model using labeled examples~\cite{devlin2018bert,sun2019fine}. 

We utilize the pool of labeled training examples for fine-tuning BERT on the task of hate speech detection, using a balanced set of hate and non-hate labeled examples (by down-sampling the latter) for this purpose. Similarly to other works, we found it beneficial to first fine-tune BERT on the related task of sentiment classification, using labeled examples from the Stanford Sentiment Treebank (SST)~\cite{socher2013recursive, wang2018glue}. We used the popular HuggingFace platform~\cite{Wolf2019HuggingFacesTS} for fine-tuning BERT, and consequently--predicting the class probabilities using the fine-tuned BERT for each generated sequence. We maintained the top-scoring 100K sequences per dataset and class--their union comprises our 1M MegaSpeech sequence corpus.  

\subsection{MegaSpeech: A Large Hate Speech resource}

We created {\it MegaSpeech}, a large-scale resource comprised of 1M text sequences, automatically generated and assessed as hate and non-hate speech.  This resource exceeds the size of currently available datasets (each comprised of thousands of examples) by magnitudes of order. We will make this resource available for research purposes upon request.

A sample of 1,000 randomly selected sequences was independently evaluated by the authors, manually assigning the generated texts to the classes of hate and non-hate.  We measured inter-annotator agreement based on a subset 250 co-annotated random examples and obtained a score of 0.712 in terms of Fleiss’ kappa~\cite{fleiss2013statistical}, indicating on strong agreement between the annotators. Overall, the ratio of generated hate sequences perceived as hate speech was as high as $65\%$, and similarly, the ratio of generated non-hate sequences perceived as non-hate was estimated at $86\%$. We further assessed the text sample with respect to coherence, or `readability' using a scale of 1--5 (ranging from poor to excellent text quality, respectively), where this evaluation yielded an average score of 3.62, i.e., overall the text quality was good. We note that the generated text resembles the seed examples genre-wise, making use of slang, abbreviations, and hashtags. Table~\ref{tab:generated_examples} includes several representative examples of hate and non-hate generated sequences. 

\begin{table}[t]
\caption{ROUGE-L scores of data generated by the GPT-2 language model per dataset. These values indicate on low similarity between the generated sequences and the sequences on which the model has been fine-tuned, for each dataset and class.}
\small
\label{tab:rouge}
\begin{tabular*}{\columnwidth}{ccc}
\toprule
     Source dataset & Generated hate & Generated non-hate\\\hline\hline

{WS} & {0.12} & {0.05}\\
{DV} & {0.07} & {0.05}\\
{FN} & {0.11} & {0.14}\\ 
{WH} & {0.09} & {0.16}\\
{SE} & {0.05} & {0.03}\\
\hline
\hline
\end{tabular*}
\end{table}

We further wished to verify that the generated sequences depart from the seed labeled examples, forming new and diverse language usage, as opposed to duplicating or repeating parts of the input examples. We therefore assessed the lexical similarity between the automatically generated sequences and the source examples per dataset and class~\cite{paraphrasing}. Concretely, we employed ROUGE-L~\cite{lin-2004-rouge}, a popular text similarity measure which indicates how much is common between two word sequences--ranging from 0 (nothing in common) to 1 (identical sequences). Table~\ref{tab:rouge} reports the computed average  ROUGE-L similarity scores. As shown, the similarity scores are low, indicating that the rich language model encoded within GPT-2 was effectively leveraged for generating new and different text sequences.

\subsection{Can MegaSpeech boost hate speech detection?}

In the following section, we report the results of an empirical study, where we train deep learning hate detection models using MegaSpeech. Let us denote those hate and non-hate text sequences in MegaSpeech that were generated given the training examples of dataset $d_i$ by $gen^i_{hate}$ and $gen^i_{non-hate}$, respectively. 
In our experiments, we augment the training examples of $d_i$ with those automatically generated examples, ideally providing hate detection classifiers that are trained using this dataset with a more general and truthful approximation of the underlying language distribution.

Our results show consistent improvements in hate detection performance by using the MegaSpeech text sequences. In particular, we show a major boost in classification {\it recall} (aka, {\it sensitivity}), which is crucial for effective hate speech detection--this impact is due to the greater diversity of hate speech introduced by MegaSpeech compared with the existing datasets.

\section{Generalized Hate Detection: Evaluation}

\begin{figure*}[t]
\centering
\includegraphics[trim={4.5cm 13cm 0cm 5cm},clip,scale=0.5]{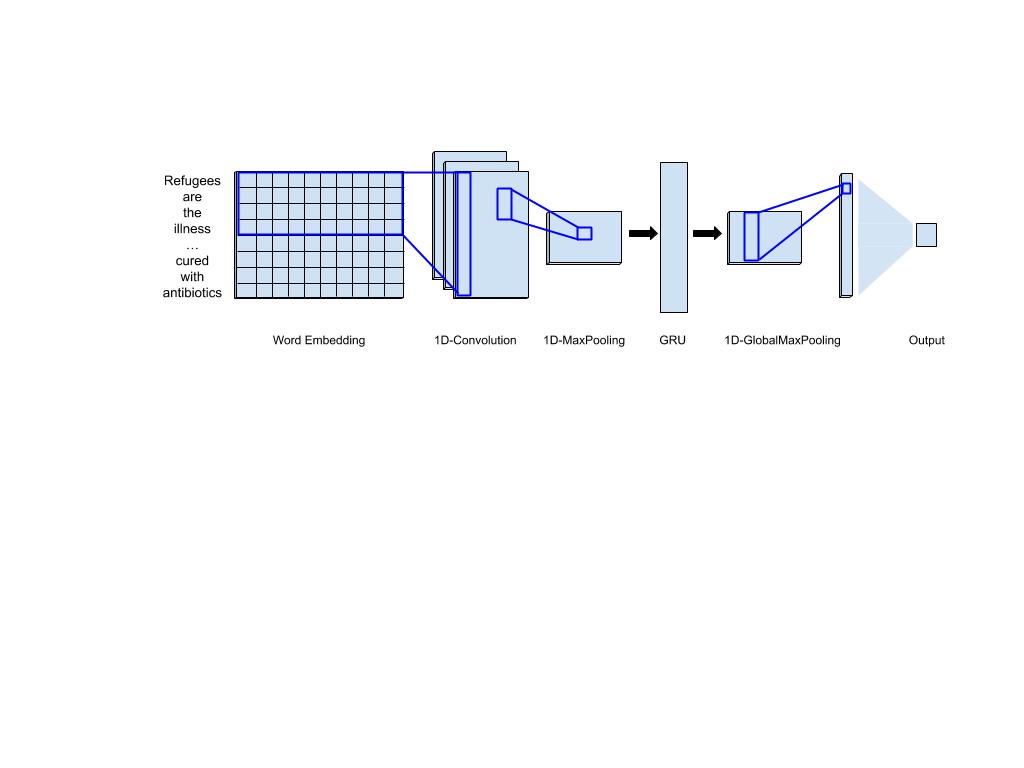}
\caption{The DL hate speech detector  \cite{zhang2018detecting}: an input text sequence is represented by a bag-of-words model, where each word is a one-hot vector. The Embedding layer reduces the dimension of each word vector from the input vocabulary dimension to 300, and the sequence of embedded word vectors are processed by a 1D convolutional layer which extracts features. The 1D MaxPooling layer down-samples the convolutional features, and the resulting down-sampled sequence enters a Gated-Recurrent Unit (GRU) which learns the time-dependencies between words. The GRU output is down-sampled by a 1D GlobalMaxPooling layer and converted to a posterior probability (that the input sequence is hate speech) by the output layer.}
\label{fig:CNN_GRU}
\end{figure*}

This section presents a comparative study of the generalization capabilities of DL-based hate speech detection, where we augment hate speech datasets with massive amounts of realistic text sequences, sampled from the approximated distribution. Concretely, each one of the datasets in Table \ref{tab:datasets} was randomly split to a training set (80\%) and a testing set (20\%). The data generation processes and all the training procedures described below were performed using the training sets, where hate detection performance is evaluated exclusively using the corresponding testing sets. We learn hate detection classification models using two types of training sets: the \emph{baseline}, which includes several thousands of human-generated text samples per dataset (Table \ref{tab:datasets}) and \emph{augmented}, constructed by adding $gen^i_{hate}$, 100,000 model-generated hate speech samples and $gen^i_{non-hate}$, 100,000 model-generated non-hate speech samples, to the corresponding baseline training set. 

Hate speech detection results were obtained using the high-performing DL-based detector \cite{zhang2018detecting}, as detailed in Figure \ref{fig:CNN_GRU}. The DL detector was implemented in TensorFlow \cite{tensorflow-2015} and trained using an NVIDIA K-80 GPU, with the Adam  \cite{Goodfellow-et-al-2016} optimizer, minimizing the binary cross-entropy loss, with early stopping and mini-batch size of 32 samples. The Convolutional layer consists of 100 filters with a window size of 4, the MaxPooling1D layer uses a pool size of 4 and the GRU outputs a 100 hidden units for each timestep. Note that in this study we compare hate detection performance given different datasets using a single DL detector, where it is not claimed that the chosen detector achieves state-of-the-art results for each of the datasets. 

Performance was evaluated using standard classification metrics: accuracy (A), precision (P), recall (R) and F1 (the harmonic mean of precision and recall). We note that the utilized datsets are generally imbalanced--and so are the derived test sets, including smaller proportions of examples labeled as hate speech vs. non-hate speech. 
We report evaluation with respect to the hate speech class: Recall corresponds to the proportion of true hate speech examples that were automatically identified as hate speech, and Precision is the proportion of correct predictions within the examples identified as hate speech by the detector. While false positive predictions may be tracked relatively easily by means of further human inspection, it is impossible to track false negatives at scale. Therefore, increasing recall by means of improved generalization is of great importance in practice.  

\subsection{Experiments}
Our study included two types of experiments, which we refer to as intra-dataset and cross-dataset hate speech detection. The {\it intra-dataset} experiments applied the DL detector to each dataset independently: performance of the DL detector was measured using the held-out test examples of every dataset $D_i$, having trained the detector with either the baseline training set, or with the augmented training set available for $D_i$. In addition, we formed a \textit{combined} dataset comprised of the union of all datasets (having joined all training and testing sets, respectively). The {\it cross-dataset} experiments aim to evaluate the realistic condition of data shift, where the hate detector is applied to new examples drawn from a data distribution that is different from the one used for training. In these experiments, having the detector trained using the training set of dataset $\mathcal{D}_i$, it was evaluated on the held-out testing set of dataset $\mathcal{D}_j\neq \mathcal{D}_i$. Again, we trained the DL detector using either the baseline or the augmented training set available for dataset $D_i$, aiming to gauge the impact of dataset augmentation on hate detection generalization; ideally, hate detection should maintain good performance when applied to new examples, where, as discussed above, it is crucial to maintain high recall during test time.

There generally exists a trade-off between precision and recall classification performances~\cite{minkov2006ner}. Given an input text sequence $\textbf{x}$, the output of the DL detector is the posterior probability $P(class(\textbf{x})="hate"|\textbf{x})$, namely, the probability that the input belongs to the "hate" class. The input is classified as "hate" if $P(class(\textbf{x})="hate"|\textbf{x})>\tau$, and otherwise as "non-hate", for some pre-defined threshold $\tau$. Different threshold choices alter the trade-off between precision and recall (with high thresholds improving precision). In order to provide a uniform view of the DL detector performance, we fixed $\tau=0.7$ \footnote{$\tau=0.7$ provided good F1 results on-average for the intra-dataset experiments.} across all experiments, although tuning $\tau$ per dataset can lead to better results for some datasets. 

\par
\begin{figure*}[ht]
\centerline{
\includegraphics[trim={1cm 0.5cm 1.25cm 0.25cm},clip,scale=1.1]{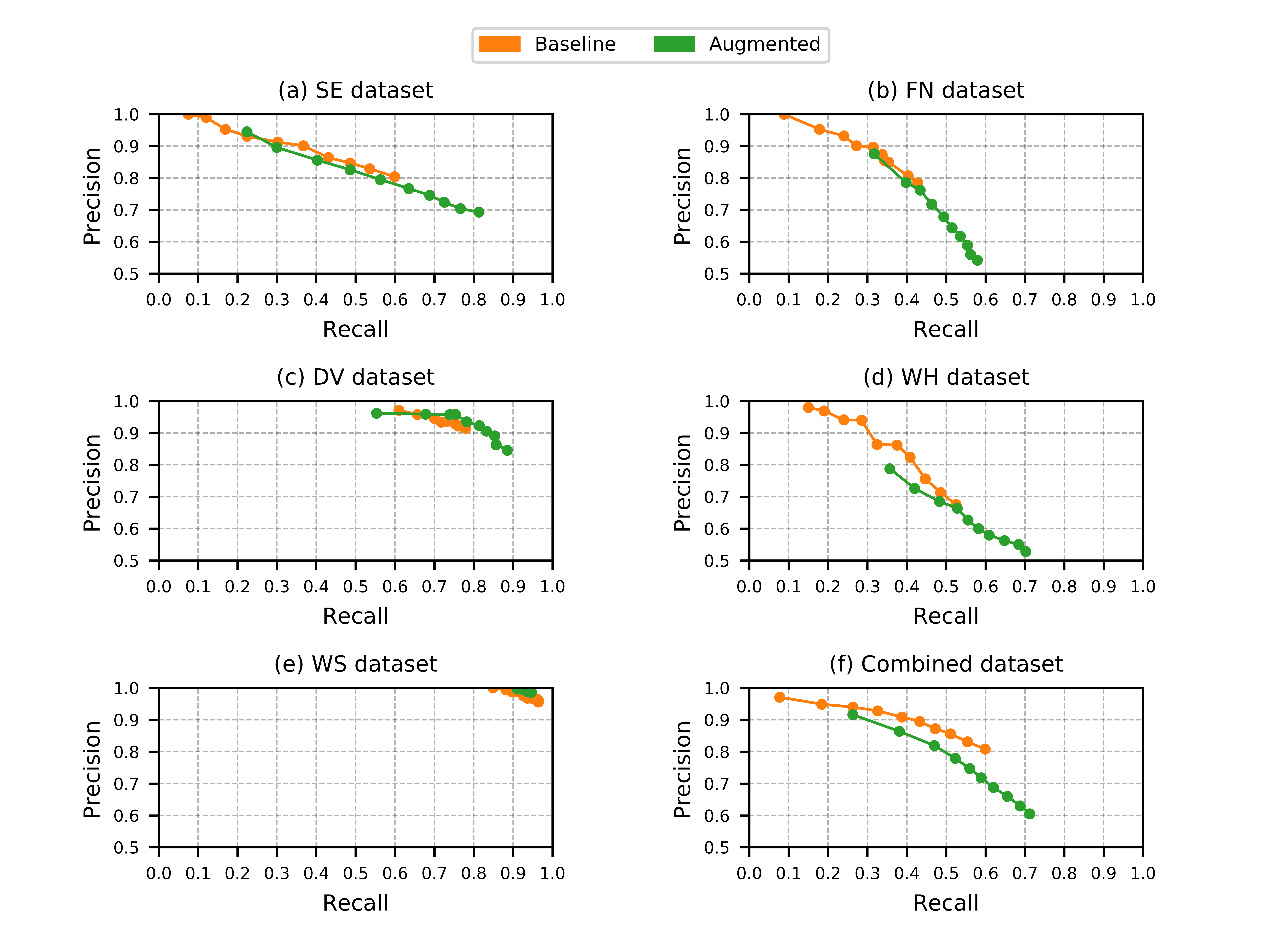}}
 \caption{Precision-Recall curves of the DL Hate Speech detector, trained separately using each dataset (a)-(e) and using the combined dataset (f), comparing baseline and augmented training data results. }
  \label{fig:PR-Curves}
\end{figure*}

 \textbf{Intra-Dataset Experiments. }
Table \ref{tab:intra_dataset_results} reports test results of the hate detector trained using the baseline (un-augmneted) verus the augmented training sets, for each of the datasets in Table~\ref{tab:datasets}, and for the combined dataset. The results indicate that augmenting the training sets with automatically generated examples leads to improvement in Recall and F1 in most cases, peaking at $+73.0\%$ and $+33.1\%$, respectively, for the \textbf{SE} dataset. Precision on the other hand is decreased for most datsets, and Accuracy changes mildly. This is not surprising, as the training and test sets belong to the same distribution and are highly similar in this setup, whereas the generated text sequences introduce language diversity, as well as some label noise. For the combined dataset, which is relatively diverse, there is a yet a boost due to train set augmentation of $+48.2\%$ in Recall and $+16.1\%$ in F1;  Precision decreases by $19.7\%$ and Accuracy almost unchanged. Therefore, by dataset augmentation, a dramatic drop in false-negatives (missed hate speech sequences) is achieved, namely, significantly more hate speech is detected, which is also evident in the improved F1. The decrease in Precision indicates more false-positives (non-hate sequences classified as hate), however, these are less severe than missed hate speech sequences. Figure \ref{fig:PR-Curves} provides the complete Precision-Recall curves for all datasets, computed by altering the detector threshold in the range of $\tau=0.5$ to 0.95 in steps of 0.05. These curves clearly demonstrates that training data augmentation yields significantly imrpved Recall (and consequently F1) across this range.

\begin{table*}[ht]
\large
\centering
\caption{Intra-Dataset classification results, comparing the baseline and augmented-baseline training sets}
\label{tab:intra_dataset_results}
\resizebox{\textwidth}{!}{
\begin{tabular}{c|ccr|ccr|ccr|ccr}
\toprule
     Dataset & \multicolumn{3}{c}{Accuracy} & \multicolumn{3}{c}{Precision} & \multicolumn{3}{c}{Recall} & \multicolumn{3}{c}{F1}\\
     & Baseline & Augmented & (\%) & Baseline & Augmented & (\%) & Baseline & Augmented & (\%) & Baseline & Augmented &(\%)\\ 
     \hline

{WS} & {0.967} & {0.977} & {+1.03} & {0.968} & {0.989} & {+2.17} & {0.936} & {0.943} & {+0.75} & {0.952} & {0.966} & {+1.47}\\\hline
{WH} & {0.891} & {0.872} & {-2.13} & {0.862} & {0.600} & {-30.40} & {0.375} & {{0.582}} & {+55.20} & {0.523} & {0.591} & {+13.00}\\\hline
{SE} & {0.715} & {0.764} & {+6.85} & {0.901} & {0.767} & {-14.87} & {0.367} & {{0.635}} & {+73.02} & {0.522} & {{0.695}} & {+33.14}\\\hline
{DV} & {0.922} & {0.935} & {+1.41} & {0.929} & {0.923} & {-0.65} & {0.753} & {0.814} & {+8.10} & {0.832} & {0.865} & {+3.97}\\\hline
{FN} & {0.956} & {0.942} & {-1.46} & {0.874} & {0.644} & {-26.32} & {0.337} & {{0.515}} & {+52.82} & {0.486} & {{0.573}} & {+17.90}\\\hline
{Combined} & {0.904} & {0.905} & {+0.11} & {0.895} & {0.718} & {-19.77} & {0.433} & {{0.642}} & {+48.26} & {0.584} & {0.678} & {+16.10}\\\hline
\end{tabular}}
\end{table*}
\par
\begin{figure*}
\centerline{
\includegraphics[trim={0cm 3.0cm 0cm 2.75cm},clip ,width=38pc]{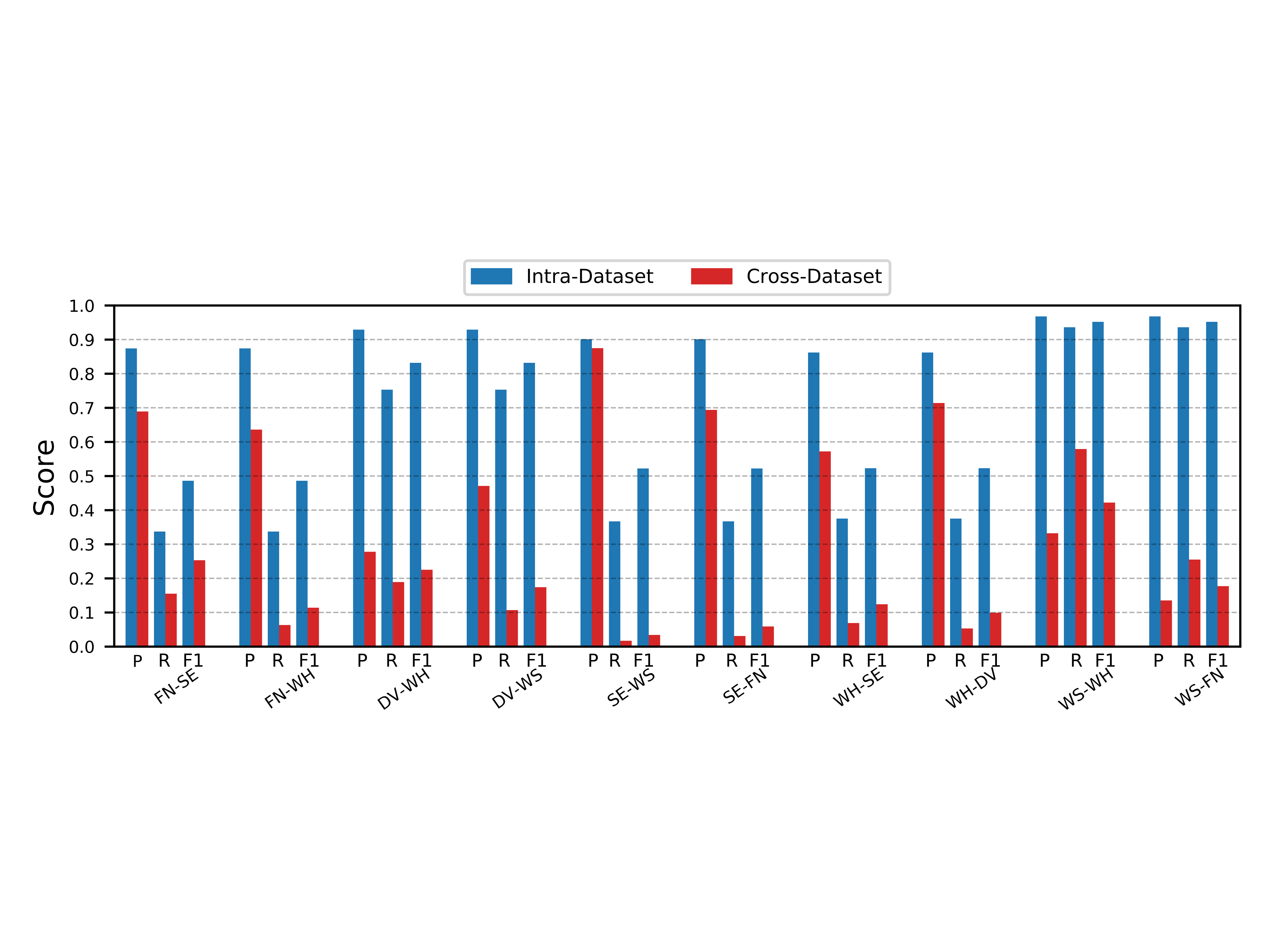}}
\caption{Cross-Dataset performance without training set augmentation: in each experiment the DL detector was trained using the training set of one dataset (the left initials of each pair). The trained detector was evaluated on the held-out test set of the same dataset (Intra-Dataset) and on the held-out test set of a different dataset (Cross-Dataset, according to the right initials of each pair). The results clearly indicate a significant decrease in cross-data performance, especially in the Recall and F1 metrics. Therefore, the proportion of false-negatives (i.e. undetected hate speech) significanlty increases, as compared to the intra-dataset results.}
\label{fig:Cross_Dataset}
\end{figure*}

 \textbf{Cross-Dataset Experiments. } 
We further evaluated the generalization of the DL detectors in conditions of data shift. Figure~\ref{fig:Cross_Dataset} shows the results of applying the DL detectors trained using the baseline (non-augmented) training set $D_i$ on the held-out test set $D_i$ (intra-dataset), as well as on the held-out test sets of other datasets, $D_j \neq D_i$ (cross-dataset). As shown, there is a significant drop in performance measurements between the first and latter setups--we observe precision drops, and poor recall performance (below 0.2) on the target test sets in conditions of data shift.
Therefore, provided with (un-augmented) train sets of few thousands of hate speech examples, the achievable generalization to other hate speech datasets is very limited, resulting in low levels of recall (i.e., high ratio of missed hate speech).

In another set of experiments, we re-trained the detector with each one of the corresponding augmented training set, and evaluated on the same cross-dataset combinations of the first experiment. Table~\ref{tab:cross_dataset} summarizes the results of the Cross-dataset experiment. The average metrics  of all cross-dataset pairs reveals a consistent increase in accuracy ($+6.9\%$), precision ($+3.5\%$), recall ($+182.8\%$) and F1 ($+179.7\%$). That is, augmenting the original labeled training sets with the automatically generated text sequences resulted in a dramatic increase of recall, while maintaining good levels of precision. Overall, this trend is reflected by a steep increase in the combined measure of F1. These results thus provide another approval of the benefits of augmenting the data with generated sentences and using a broader data distribution by utilizing a pre-trained language model. 

\begin{table*}[h]
\large
\centering
\caption{Cross-Dataset performance, comparing detector training using baseline to augmented-baseline training sets}

\label{tab:cross_dataset}
\resizebox{\textwidth}{!}{

\begin{tabular}{l|ccr|ccr|ccr|ccr}

\toprule
     Trainset-  &  \multicolumn{3}{c}{Accuracy} & \multicolumn{3}{c}{Precision} & \multicolumn{3}{c}{Recall} & \multicolumn{3}{c}{F1} \\ 

     Testset   & Baseline & Augmented & (\%) & Baseline & Augmented & (\%) & Baseline & Augmented & (\%) & Baseline & Augmented &(\%)\\\hline

{FN-SE} & {0.613} & {0.645} & {+5.22} & {0.689} & {0.570} & {-17.27} & {0.155} & {0.644} & {+315.48} & {0.253} & {0.605} & {+139.13}\\ \hline
{FN-WH} & {0.846} & {0.850} & {+0.47} & {0.636} & {0.528} & {-16.98} & {0.063} & {0.507} & {+704.76} & {0.114} & {0.517} & {+353.50}\\ \hline
{DV-WH} & {0.794} & {0.820} & {+3.27} & {0.278} & {0.441} & {+58.63} & {0.189} & {0.507} & {+168.25} & {0.225} & {0.472} & {+109.77}\\ \hline
{DV-WS} & {0.652} & {0.832} & {+27.60} & {0.471} & {0.870} & {+84.71} & {0.107} & {0.599} & {+459.81} & {0.174} & {0.709} & {+307.47}\\ \hline
{SE-WS} & {0.662} & {0.754} & {+13.89} & {0.875} & {0.872} & {-0.34} & {0.017} & {0.331} & {+1,847.06} & {0.034} & {0.480} & {+1,311.76}\\ \hline
{SE-FN} & {0.926} & {0.924} & {-0.21} & {0.694} & {0.489} & {-29.53} & {0.031} & {0.226} & {+629.03} & {0.059} & {0.309} & {+423.72}\\ \hline
{WH-SE} & {0.584} & {0.593} & {+1.54} & {0.572} & {0.522} & {-8.74} & {0.069} & {0.450} & {+552.17} & {0.124} & {0.484} & {+290.32}\\ \hline
{WH-DV} & {0.752} & {0.777} & {+3.32} & {0.714} & {0.610} & {-14.56} & {0.053} & {0.364} & {+586.79} & {0.099} & {0.456} & {+360.60}\\ \hline
{WS-WH} & {0.749} & {0.824} & {+10.01} & {0.332} & {0.450} & {+35.54} & {0.579} & {0.489} & {-15.54} & {0.422} & {0.469} & {+11.13}\\ \hline
{WS-FN} & {0.822} & {0.895} & {+8.88} & {0.135} & {0.233} & {+72.59} & {0.255} & {0.176} & {-30.98} & {0.177} & {0.201} & {+13.55}\\ \hline\hline
Average & 0.740	  & 0.791	& \textbf{+6.95}	& 0.539	  & 0.559	 & \textbf{+3.50}	& 0.151   & 0.429	& \textbf{+182.81}	& 0.168	& 0.470& 	\textbf{+179.71}\\ \hline
\end{tabular}}
\end{table*}

\begin{table*}
\caption{Examples of generated sequences by the GPT-2 language model, fine-tuned by each source training set.}
\label{tab:generated_examples}
\vskip 5pt
\resizebox{\textwidth}{!}{
\small
\centering
\begin{tabular}{c|p{6cm}|p{6cm}}
\hline
\textbf{Dataset} & \textbf{Hate Speech Generated by GPT-2} & \textbf{Non-Hate Speech Generated by GPT-2}\\
\hline
\multirow{ 2}{*}{WS} & What does Islam have to
do with democracy and human rights islam is a hateful religion that has inspired the murders of & The haters are all over this comment\\\cline{2-3}
& for starters i know some of you never read this canadian quran is full of excuses why so many muslims are murdering islamists & This story originally appeared on National Geographic News\\\cline{2-3}
& the writer of the sharia law book given to the muslim world in the 1980s is calling for a religious civil war & US wants Zambias political crisis to end \\
\hline
\multirow{ 2}{*}{DV} & \#AltRight is so important to white people these days they want to rebrand white genocide as immigration amnesty \#WhiteGenocide \#NoMoreD & A member of the Liberty Bell Choir sings the National Anthem before the game between the New York Yankees \\\cline{2-3}
& Fox News allowed these other nut jobs on their network to spout hatred towards women and minorities so that they could sell more toys to these scum pigs & Who do you think of when you think of one of sports lowest paid employees in league history \#yankees \#mariners \#mlb\\\cline{2-3}
& we must stop lawless illegal black hearted sand nigger lovers theyre ruining our country my countrymen dont like those people & Afghanistans military training looks to be on the brink of a revolution after President Ashraf Ghani pledged to make it so special that it is\\
\hline
\multirow{ 2}{*}{FN} & this new administration is willing to start a world war over corruption while these illegal immigrants destroying our country are not taken into consideration & On Thursday evening a collection of posts appeared on a website with the slogan stop the racist terrorist threats to burma \#facebook \#burmaco\\\cline{2-3}
& i wish someone would tell me what the hell is going on with this video i hate it man that fucking nigga & what the fuck does feeling good imply hes not feeling good anymore ill shake his hand or give him a hug mfw\\\cline{2-3}
& Democrats and their media sycophants do not want to talk about why illegal border crossers are stealing from citizens who pay taxes \#buildthatwall & It was initially deemed a case of falling victim to men in power who are trying to change the \#innovation \#technology\\
\hline
\multirow{ 2}{*}{WH} & The number of refugees and immigration controls are clear, now is it time to go to extremes in a desperate attempt to stop them & thanks for your reponse this is what our school system is becoming \\\cline{2-3}
& the left celebrates the death of freedom as it is the road to the extermination of whites and the rebirth of the baboon & as an asian guy living in canada i have to say i love it here welcome to stormfront\\\cline{2-3}
& Mexicans wonder why their country is going to hell when they send their sons and daughters to fight for such disgusting filth & i remember when my 3 year old son made a face like the one on the far left in the above cartoon \\
\hline
\multirow{ 2}{*}{SE} & Afghan refugees now run our towns and streets \#buildthatwall \#secureourborders \#unleashtheinsane & Police say a man in his 20s is dead and three others were injured after a brawl broke out at a political rally in Athens\\\cline{2-3}
& illegal aliens must be deported or even sued if not they will keep coming illegally \#nodaca \#noamnesty \#endbirthrightcitizenship & To celebrate the launch of the Fund for Refugees I am inviting you to become a guardian of this important right for every Syrian refugee as we defend refugee rights\\\cline{2-3}
& Hillary Clinton Confronts Gay Muslim Refugee \#Rapefugee who bragged about the way he raped a little boy & Republican presidential candidate Donald Trump has vowed to ensure anything is possible with all of our great employees \#muslimimmigration partners \#bayareamus\\
\hline
\end{tabular}}
\end{table*}

\section{CONCLUSIONS}
In this work we demonstrated the problems that may occur when training a hate speech classifier on small datasets. We showed the performance decrease that arisen when training a hate speech classifier on a small dataset and tested on a dataset of a different hate speech language distribution. We suggested a new framework for augmenting hate speech dataset by using data that was generated from a deep generative model trained on the small hate speech datasets that were available. The proposed framework was evaluated using a high-performing DL hate speech detector, demonstrating a significant boost in the generalization capability of the DL detector, intra-dataset and cross-dataset as well.

\subsection{ACKNOWLEDGMENT}

This research was supported partly by Facebook Content Policy Research on Social Media Platforms Research Award.

\bibliographystyle{IEEEtran}
\bibliography{references}

\end{document}